\pdfoutput=1
\documentclass[11pt]{article}

\usepackage{emnlp2021}

\usepackage{times}
\usepackage{bm}
\usepackage{amsmath}
\usepackage{amsfonts}
\usepackage{graphicx}
\usepackage{latexsym}
\usepackage{multirow}
\usepackage{arydshln}

\usepackage{hyperref}
\hypersetup{hypertex=true,
            colorlinks=true,
            linkcolor=blue,
            anchorcolor=blue,
            citecolor=blue}

\usepackage[T1]{fontenc}
\usepackage[utf8]{inputenc}

\usepackage{microtype}

\title{Exploiting Curriculum Learning in Unsupervised Neural Machine Translation}

\author{Jinliang Lu$^{1,2}$ \and Jiajun Zhang$^{1,2}$ \thanks{\ \  Corresponding author} \\
  $^{1}$National Laboratory of Pattern Recognition, Institute of Automation, CAS \\
  $^{2}$School of Artificial Intelligence, University of Chinese Academy of Sciences \\
  \texttt{\{jinliang.lu, jjzhang\}@nlpr.ia.ac.cn} \\}

\begin{document}
\maketitle
\begin{abstract}

% This document is a supplement to the general instructions for *ACL authors. It contains instructions for using the \LaTeX{} style files for EMNLP 2021. 
% The document itself conforms to its own specifications, and is therefore an example of what your manuscript should look like.
% These instructions should be used both for papers submitted for review and for final versions of accepted papers.

% Unsupervised neural machine translation (UNMT) has recently achieved significant improvements.

Back-translation (BT) has become one of the de facto components in unsupervised neural machine translation (UNMT), and it explicitly makes UNMT have translation ability. However, all the pseudo bi-texts generated by BT are treated equally as clean data during optimization without considering the quality diversity, leading to slow convergence and limited translation performance. To address this problem, we propose a curriculum learning method to gradually utilize pseudo bi-texts based on their quality from multiple granularities. Specifically, we first apply cross-lingual word embedding to calculate the potential translation difficulty (quality) for the monolingual sentences. Then, the sentences are fed into UNMT from easy to hard batch by batch. Furthermore, considering the quality of sentences/tokens in a particular batch are also diverse, we further adopt the model itself to calculate the fine-grained quality scores, which are served as learning factors to balance the contributions of different parts when computing loss and encourage the UNMT model to focus on pseudo data with higher quality.  Experimental results on WMT 14 En$\leftrightarrow$Fr, WMT 16 En$\leftrightarrow$De, WMT 16 En$\leftrightarrow$Ro, and LDC En$\leftrightarrow$Zh translation tasks demonstrate that the proposed method achieves consistent improvements with faster convergence speed.\footnote{Our code is available in \texttt{\url{https://github.com/JinliangLu96/CL\_UNMT}}}

\end{abstract}

\section{Introduction}

Unsupervised neural machine translation (UNMT) \cite{artetxe2018unsupervised,lample2018unsupervised} has made significant progress \cite{NEURIPS2019_c04c19c2,song2019mass,liu-etal-2020-multilingual,NEURIPS2020_1763ea5a} in recent years. It consists of three main components: the initialization of the cross-lingual pre-trained language model (PLM), denoising auto-encoder (AE) \cite{10.1145/1390156.1390294}, and back-translation (BT) \cite{sennrich-etal-2016-improving}. BT generates pseudo bi-texts for training and explicitly enables its translation ability. However, pseudo bi-texts are quite diverse in quality, and the low-quality bi-texts are difficult to learn. Equally treating pseudo bi-texts as clean data would negatively influence the convergence process and harm the translation performance \cite{fadaee-monz-2018-back}.

% confuse the model when optimizing, leading to the slow convergence speed and limited translation performance. 

% usually: 1) consumes extra time when producing pseudo bi-text ; 3) converges slowly due to the uneven quality of pseudo bi-text, leading to the training process less efficient. 

% ignoring the quality of pseudo bi-text in BT steps brings great noise into training, leading to the process inefficient.

% However, UNMT usually: 1) needs a massive amount of monolingual data for training; 2) consumes extra time when producing pseudo bi-text; 3) converges slowly due to the uneven quality of pseudo bi-text. These peculiarities make UNMT expensive to train.

% helps UNMT achieve remarkable results on similar language pairs .

\begin{figure}
    \centering
    \includegraphics[scale=0.23]{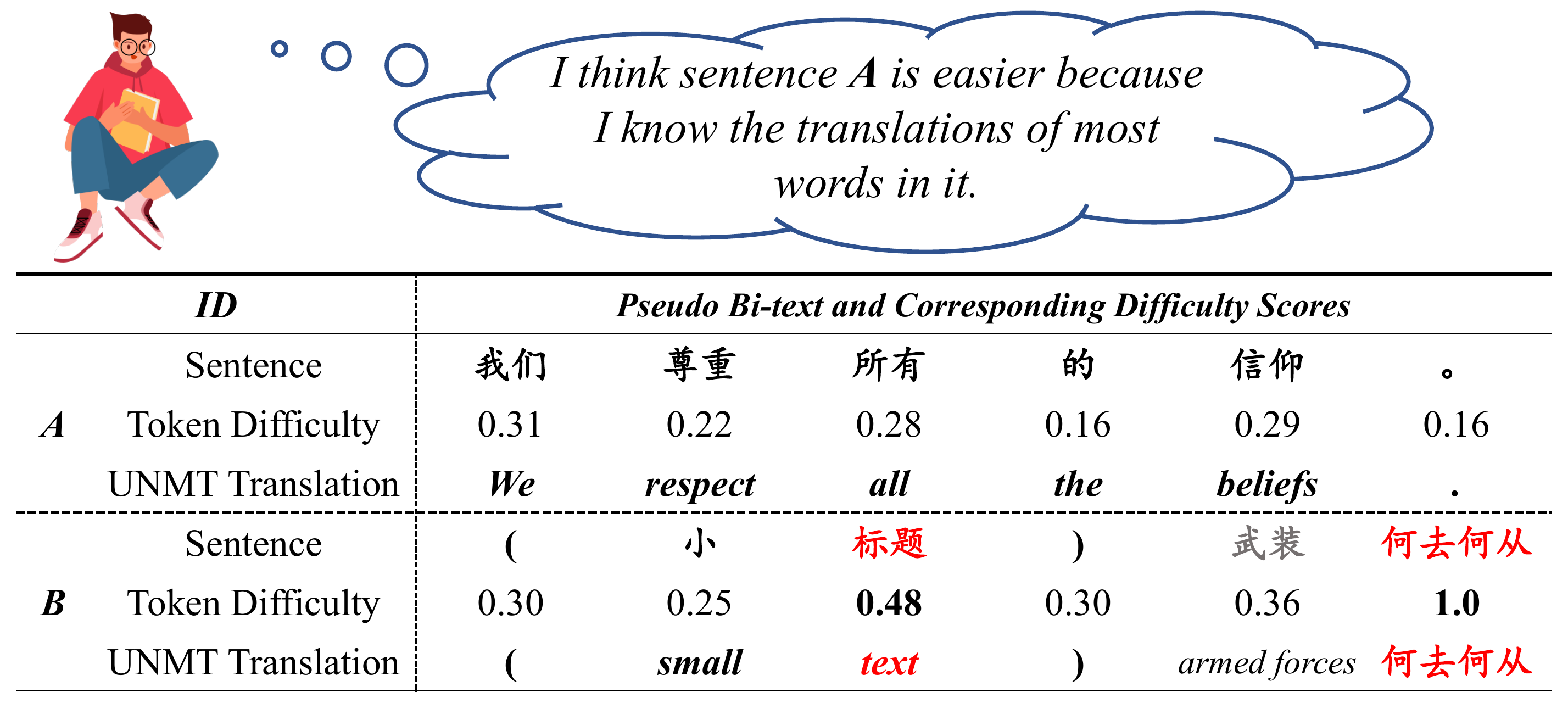}
    \caption{Difficulty scores in \textit{A} are lower than \textit{B}. And its translation is credible, making pseudo bi-text \textit{A} better (red words in \textit{B} are mis-translated or untranslated).} 
    \label{fig:1}
\end{figure}

% However, current UNMT systems usually need a massive amount of monolingual data for training. Producing pseudo bi-text at each step further consumes extra time, leading to the training process expensive.

% Except the initialization of PLM, auto-encoder (AE) \cite{10.1145/1390156.1390294} and back-translation(BT) \cite{sennrich-etal-2016-improving} also make significant contributions for UNMT. However, most works concentrate on the improvements of the cross-lingual ability of PLM, ignoring the possibility of AE and BT.
Recently, curriculum learning (CL) \cite{10.1145/1553374.1553380}, which aims to help the model learn from easy samples to the hard ones, has shown its effectiveness in speeding up the convergence and improving performance. Just as the name implies, the critical point of CL is \textit{difficulty criteria}. \citet{zhang2018empirical} classify criteria in supervised machine translation into linguistic-inspired criteria \cite{kocmi-bojar-2017-curriculum} and model-based criteria \cite{zhang-etal-2017-boosting,zhang-etal-2019-curriculum,zhou-etal-2020-uncertainty,xu-etal-2020-dynamic}. Most of them are designed from the perspective of the source side in the pure parallel corpus. However, pseudo bi-texts produced by BT with monolingual sentences in UNMT contain different levels of noise, and low-quality samples with much noise would be difficult for the model to learn appropriately \cite{guo2018curriculumnet,zhang2020self}. In this paper, we propose a CL method to gradually utilize pseudo bi-texts for UNMT from easy to hard, helping the model concentrating on the data with high quality from multiple granularities.

% Linguistic-inspired difficulty criteria, such as sentence length or word rarity, are easy to access\cite{kocmi-bojar-2017-curriculum}. Model-based difficulty criteria are usually established on the existing models(e.g., PLMs, models saved in previous steps)\cite{zhou-etal-2020-uncertainty,zhang-etal-2017-boosting,zhang-etal-2019-curriculum} or the current one\cite{xu-etal-2020-dynamic}. Through utilizing model-uncertainty\cite{gal2016dropout} or gradient information, model-based difficulty could be more suitable for training.

Intuitively, pseudo bi-text with high quality is more accessible and suitable for UNMT optimization. Accordingly, we will measure the sample difficulty with bi-text quality. First, we apply the unsupervised cross-lingual word embedding \cite{lample2018word} to calculate the quality of bi-texts, which is in turn used to measure the sample quality. Then, samples will be fed into UNMT from easy to hard batch by batch based on their difficulty. Figure \ref{fig:1} illustrates that it is reasonable to bridge the bi-text quality and the sample difficulty. 

% translation generated from the simple sentence is of higher translation confidence. Therefore, pseudo bi-text produced by them is with less noise and more suitable for optimization. Intuitively, simple sentences usually contain more words that are easy to translate.
% As shown in Figure \ref{fig:1}, humans can remember word translation pairs in the brain with different levels of assurance. Sentences with familiar words can be translated with higher confidence. We assume the model performs like us. 

However, the batch-based standard learning procedure is coarse-grained, and the qualities of pseudo bi-texts at sentence/word-level in a particular batch are also different, which should be addressed. To perform such fine-grained learning from easy to difficult, we borrow the idea from self-paced learning \cite{NIPS2010_e57c6b95}, which is an adapted CL algorithm. Specifically, we first adopt the model to estimate the quality scores of pseudo bi-texts. Then, the scores are served as learning factors to balance the contributions of different parts when computing the training loss, encouraging the UNMT model to concentrate on the parts with higher quality.

In general, the contributions of this paper can be summarized as follows:
\begin{enumerate}
    \item[$\bullet$] We propose a multi-granularity CL method to improve UNMT. To the best of our knowledge, this is the first attempt to study the CL framework for UNMT.
    \item[$\bullet$] Through utilizing the quality of pseudo bi-text from multi-granularities, our method helps UNMT concentrate on the easy-to-learn part of data and optimize in the proper direction.
    \item[$\bullet$] Extensive experiments on WMT14 En$\leftrightarrow$Fr, WMT16 En$\leftrightarrow$De, WMT16 En$\leftrightarrow$Ro, and LDC En$\leftrightarrow$Zh translation tasks demonstrate that our method consistently outperforms the strong baselines with faster convergence speed.
\end{enumerate}

% Furthermore, we find that CL strategy is better for distant language pair like en$\leftrightarrow$zh, while SPL strategy is more suitable for the similar language pairs. The combination of the two strategies reveals the best results.

% Specifically, we make the following contributions in this paper:
% \begin{enumerate}
%     \item We propose cross-lingual enhanced CL for UNMT, 
% \end{enumerate}

\section{Background of UNMT}
The architecture of the current state-of-the-art UNMT is the same as supervised NMT model, except that the UNMT model simultaneously processes both translation directions. The training procedure comprises three main components: the initialization of cross-lingual PLM, denoising auto-encoder and back-translation. 

\textbf{Cross-lingual PLM} is the auto-encoder that aims to encode the source sentences and target sentences into a shared embedding space. The parameters are used to initialize the encoder and decoder in UNMT model before training.

\textbf{Denoising Auto-Encoder} is one of the crucial components for UNMT. It can improve the model learning ability through reconstructing the original sentences from the sentences with artificial noise, such as random deletion, swapping, or blanking. It is optimized by minimizing the following objective function:
\begin{equation}
    \begin{split}
        \mathcal{L}_{auto} &= \mathbb{E}_{x \sim \phi_{l_{1}}}\left[-\log P_{l_{1} \rightarrow l_{1}}(x|C(x))\right] \\ 
        & + \mathbb{E}_{y \sim \phi_{l_{2}}}\left[-\log P_{l_{2} \rightarrow l_{2}}(y|C(y))\right]
    \end{split}
\end{equation}
where $x$ and $y$ indicate sentences sampled from monolingual dataset $\phi_{l_{1}}$ and $\phi_{l_{2}}$. $l_{1}$ and $l_{2}$ are the two languages. $C(\cdot)$ is the artificial noise function.

\textbf{Back Translation} is another essential component of UNMT, which explicitly ensure the model to have translation ability. First, each batch of monolingual sentences is translated into the other language by UNMT model $M$. Then, $M$ applies the pseudo parallel sentences $(M_{{l}_{1} \rightarrow l_{2}}(x), x)$ and $(M_{{l}_{2} \rightarrow l_{1}}(y), y)$ into training. The process is called \textit{on-the-fly back translation}. The objective function is:
\begin{equation}
    \begin{split}
        \mathcal{L}_{bt} &= \mathbb{E}_{x \sim \phi_{l_{1}}} \left[-\log P_{l_{2} \rightarrow l_{1}} (x|M_{{l}_{1} \rightarrow l_{2}}(x)) \right] \\
        & + \mathbb{E}_{y \sim \phi_{l_{2}}} \left[-\log P_{l_{1} \rightarrow l_{2}} (y|M_{{l}_{2} \rightarrow l_{1}}(y))\right]
    \end{split}
\end{equation}

In conclusion, the final loss during UNMT training can be written as follow:
\begin{equation}
    \mathcal{L} = \mathcal{L}_{auto} + \mathcal{L}_{bt}
\end{equation}

Even though strong UNMT models have been proposed in recent years, such as XLM \cite{NEURIPS2019_c04c19c2} and MASS \cite{song2019mass}. The uneven quality of pseudo bi-text is still harmful. First, pseudo bi-texts are produced at each round. The translation performance in the early stages is pretty low and will affect the final results. Second, equally treating pseudo bi-texts with uneven quality can bring deviation to the optimization, slowing down the convergence speed and restricting translation performance.

\section{Approach}
\begin{figure}
    \centering
   	\scalebox{0.38}{\includegraphics{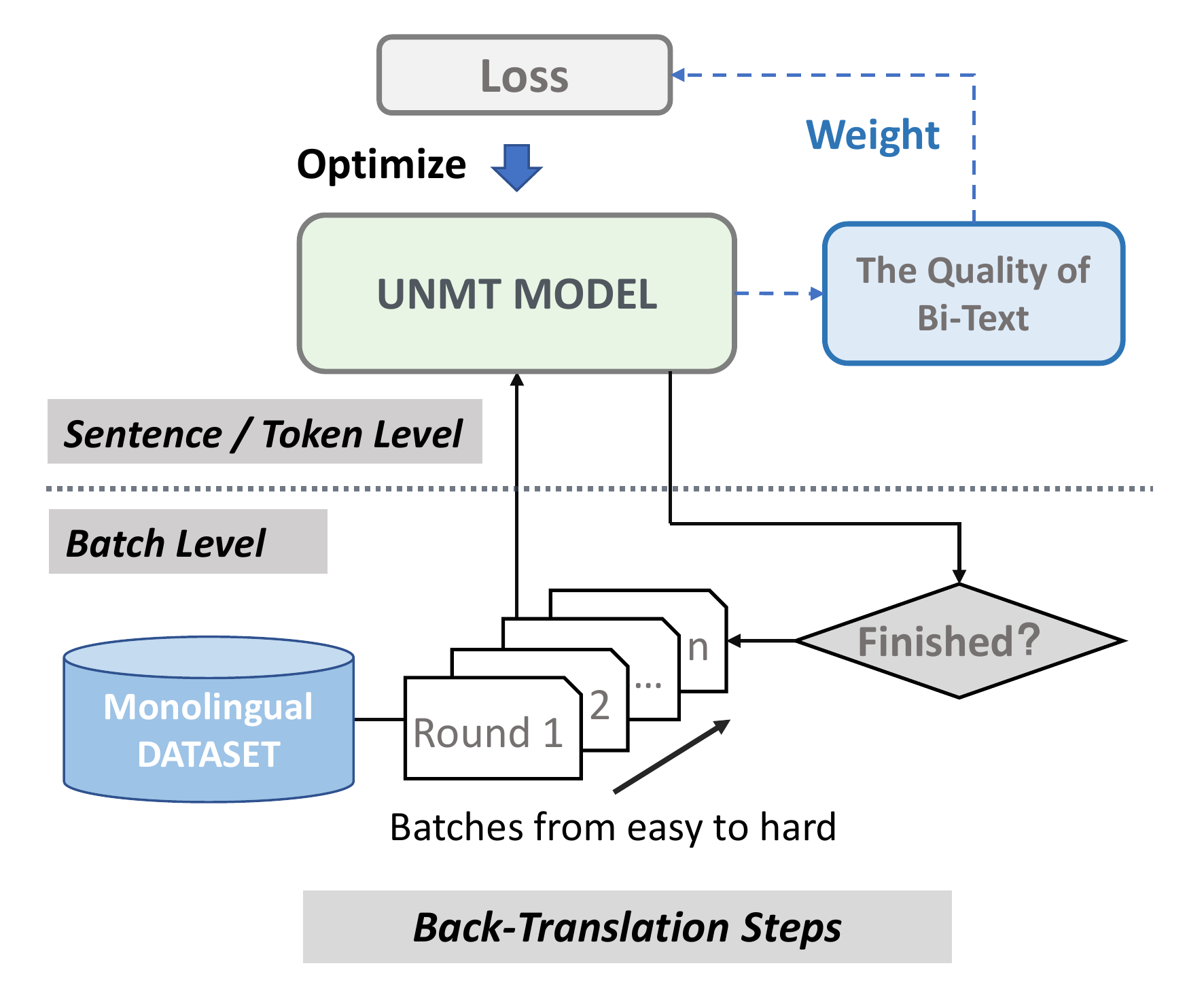}}
  	\caption{Illustration of our method. Batch level CL is shown below the black dash line, which controls the dataloader to prepare batches based on sample difficulty. Sentence/token level CL is illustrated above the black dash line, applying UNMT model to estimate the quality of pseudo bi-text and weight the training loss.}
  	\label{fig.2}
\end{figure}

In this section, we introduce the proposed CL method for UNMT. As shown in Figure \ref{fig.2}, our method consists of two sub-modules that work at different levels:
\begin{enumerate}
    \item[1)] At batch level, we aim to optimize the dataloader so as to load the samples for training from easy to difficult batch by batch (§ \ref{sec.3.1});
    \item[2)] At sentence/token level, we attempt to improve the parameter optimization procedure by using an adapted CL algorithm self-pace learning \cite{NIPS2010_e57c6b95}, which calculates fine-grained difficulty scores and encourages the optimizer to pay more attention on easy-to-learn sentences/tokens (§ \ref{sec.3.2}). 
\end{enumerate}

% 1) Utilize  traditional CL methods with a noval cross-lingual difficulty criterion to control the dataloader, gradually increasing the difficulty of samples at the batch level (§ \ref{sec.3.1}); 2) Adopt self-paced learning \cite{NIPS2010_e57c6b95}, an adapted CL algorithm, to consider the difficulty in fine-trained levels. Specifically, UNMT is employed to estimate the quality of pseudo bi-text at the sentence level and token level, automatically regulating its training loss and then directing the optimization (§ \ref{sec.3.2}).

% As shown in Figure \ref{fig.2}, cross-lingual difficulty enhance CL controls the dataloader, gradually increasing the difficulty batch by batch. And fine-grained translation confidences calculated by the model itself are applied to weight the loss function, automatically directing the optimization.

% Then, we separately introduce the sub-modules.

\subsection{Batch Level CL}
\label{sec.3.1}
In this section, we introduce the CL method which controls the dataloader to load samples from easy to hard at the batch level. First, we describe the cross-lingual difficulty definition for the measurement of training samples. Then, we explain the sample loading schedule for UNMT.

% with novel cross-lingual difficulty criterion,  which first introduce the cross-lingual based difficulty criterion for the measurement of training samples. Then, we describe the modified learning schedule that determines how difficult samples can be used during the training process. 

% that helps UNMT model gradually learn from easy to hard.

% Difficulty criteria and learning schedule are the main components in CL. 

\subsubsection{Difficulty Criterion Definition}
\label{sec.3.1.1}
As mentioned above, difficulty criterion is essential for CL. Traditional criteria, such as sentence length or word rarity, cannot reflect the practical complexity of pseudo bi-text. 

% difficulty definition in UNMT is tough because the quality of bi-text constructed in BT steps cannot be guaranteed. 

% To address this problem, we use cross-lingual distance to depict the difficulty of samples.

We first use cross-lingual similarity to calculate the word-level bi-text quality, which is in turn used to define the word-level difficulty.
%Based on the assumption that easy words are less likely to be mistranslated, the quality of pseudo source sentences translated by texts with more easy words could be better. Therefore, we first estimate the translation difficulty of a specific word by the shortest distance between it and the target language space.
Then, we weight the word-level difficulties by importance to get the sentence-level difficulty.

% Since UNMT always apply great amount of monolingual sentences into training, we choose static difficulty for the consideration of speed. To address the importance of bilingual information which is ignored by prior works, we take the cross-lingual similarity into account.

Specifically, pre-trained monolingual word embedding of language $X$ and $Y$ are first mapped into the same latent space through MUSE \cite{lample2018word} toolkit and cross-lingual embedding matrices $\bm{Z}_{X}, \bm{Z}_{Y}$ are obtained. Next, sentence $x^{i} = \left<x_{1}^{i}, x_{2}^{i}, \cdots, x_{n}^{i}\right>$ is mapped into a sequence of vectors $\bm{x}^{i} = \left[\bm{x}_{1}^{i}, \bm{x}_{2}^{i}, \cdots, \bm{x}_{n}^{i}\right]$ through $\bm{Z}_{X}$. Then, the difficulty of word $x_{j}^{i}$ can be calculated, which is represented by the shortest distance from it to the target language space $\bm{Z}_{Y}$:
\begin{gather}
    d(x_{j}^{i}) = 1 - \max \limits_{\bm{z}_{k} \in \bm{Z}_{Y}} \cos(\bm{x}_{j}^{i}, \bm{z}_{k})
\end{gather}
where $\bm{z}_{k}$ indicates an arbitrary word embedding in $\bm{Z}_{Y}$. Considering the contribution of different words, sentence-level difficulty calculation incorporates importance weighting (indicated by \texttt{tfidf} score). Sentence length is further applied as the penalty. To sum up, the formula can be written as:
\begin{gather}
    d(x^{i}) = \frac{\sum_{j=1}^{n} \text{tfidf}(x_{j}^{i}) \cdot d(x_{j}^{i})}{\sum_{j=1}^{n} \text{tfidf}(x_{j}^{i})} \cdot \log(n)
\end{gather}

Finally, the difficulties are normalized to [0, 1] by minmax normalization, employed during the batch preparation.

\subsubsection{Sample Loading Schedule}
The second question in CL is how to design the sample loading schedule, which determines how complex samples the UNMT can accept at specific steps. We follow the competence definition designed by \citet{platanios-etal-2019-competence}, which indicates the capacity of the model:
\begin{gather}
    c(t) = \min (1, \sqrt[p]{\frac{t}{T} (1 - c_{0}^{p}) + c_{0}^{p}})
    \label{eqa:6}
\end{gather}
where $c_{0}$ is the initial competence, $p$ (set as 2 in our experiments) is the coefficient to control the curriculum schedule. At step $t$, sentences with $d(x^{i}) \leq c(t)$ become accessible to the model. And $T$ determines the step when all the sentences become available for the model. 

We compute UNMT competence at specific steps for efficiency. At the beginning of the training process, all the available samples ($d(x^{i}) \leq c_{0}$) are grouped into batches. Then, the batches are shuffled and successively transported into the model. When all of them are used up, the next phase will start with the update of $c_{t}$, sentence selection, and the batch preparation. Through the learning schedule, UNMT gradually receives the samples from easy to hard batch by batch.

% However, sampling a batch at each iteration can gradually slow down the training speed. On the other hand, uniform sampling may force the model to concentrate overly on the samples already used, which may harm the performance of UNMT.

% To address this problem, we compute UNMT competence at specific steps. According to the initial competence $c_{0}$, all the available samples ($d(x^{i}) \leq c_{0}$) are grouped into batches. Then, the batches are shuffled and successively transported into the model. When all of them are used up, the next phase will start with the update of $c_{t}$ and the batch preparation.

\subsection{Sentence/Token Level CL}
\label{sec.3.2}
While batch level CL controls dataloader to help UNMT learn from easy samples to the hard ones gradually, the qualities and difficulties of sentences and words in a particular batch are also different. However, such fine-grained operations are not suitable for the dataloader. To address this problem, we apply the UNMT to estimate the quality of pseudo bi-text at sentence/word level. Then, the quality scores are employed to regulate training loss, helping UNMT automatically focus on the words and sentences with high quality.

% Therefore, obtaining accurate confidence estimation may prevent the model from learning incorrect alignments and improving performance.

%  \cite{blatz-etal-2004-confidence, specia2009estimating}

% To address the problem, \cite{wan-etal-2020-self} propose self-paced learning for supervised NMT, applying uncertainty-based confidence to automatically weight loss function so that the model can guide itself. Specifically, the definition of model uncertainty can be written as
% \begin{gather}
%     u = \mathcal{F} \{P(t|s;\bm{\theta^{k}})\}_{k=1}^{K}
% \end{gather}
% Where $\mathcal{F}$ represents variance Var or expectation $\mathbb{E}$ which need $K$-Pass forward computation with different dropout. Obviously, Bigger \textit{K} brings not only accurate estimation but also time consumption.

% Unlike supervised NMT, the crucial problem in UNMT is the quality of pseudo source sentences, which can also be relatively viewed as the translation confidence on the target side. Therefore, obtaining accurate quality estimation\cite{blatz-etal-2004-confidence, specia2009estimating} using $\bm{\theta}$ may prevent the model from learning incorrect alignments and improving performance.

\subsubsection{Cross-lingual PLM based Pseudo Bi-text Quality Estimation (CP)}
\label{sec.3.2.1}
Cross-lingual PLMs are proven to be effective on reference-free evaluation in machine translation \cite{qi-2019-nju,yankovskaya-etal-2019-quality,kim-etal-2019-qe,zhao-etal-2020-limitations,takahashi-etal-2020-automatic}. Actually, the encoder of UNMT, which is initialized by the cross-lingual PLM, should also be able to judge the quality of pseudo bi-texts. Furthermore, since sentences/token level CL works on optimization, we apply the model to calculate dynamic quality estimation related to the current learning state instead of utilizing static scores.

Specifically, sentence $x^{i}=\left<x_{1}^{i}, x_{2}^{i}, \cdots, x_{n}^{i}\right>$ is sampled from the monolingual dataset, its \textit{on-the-fly} translation is $\hat{y}^{i}=\left<\hat{y}_{1}^{i}, \hat{y}_{2}^{i}, \cdots, \hat{y}_{m}^{i}\right>$. We apply the encoder to obtain the hidden states $\bm{H}_{x_{i}} = [ \bm{h}_{x_{1}^{i}}, \bm{h}_{x_{2}^{i}}, \cdots, \bm{h}_{x_{n}^{i}} ]$ and $\bm{H}_{\hat{y}^{i}} = [ \bm{h}_{\hat{y}_{1}^{i}}, \bm{h}_{\hat{y}_{2}^{i}}, \cdots, \bm{h}_{\hat{y}_{m}^{i}} ]$. Then, the hidden states are employed to estimate the quality of pseudo bi-text.
% Intuitively, the quality of translation can be measured by the cosine-similarity on the hidden states.

\noindent \textbf{Token-Level Translation Quality (TTQ)}: For token $x_{j}^{i}$, we use the greedy matching strategy to match it to the most similar token in $\hat{y}^{i}$. The corresponding quality of $x_{j}^{i}$ is represented by the cosine similarity, which can be formulated as:
\begin{gather}
    w = \max \limits_{v \in {1, 2, \cdots, m}} \cos(\bm{h}_{x_{j}^{i}}, \bm{h}_{y_{v}^{i}}) \\
    \hat{\alpha}_{j}^{i} = w ^ {k}
    \label{eqa:7}
\end{gather}

where $k$ is hyper-parameter for the quality gap scaling. To stabilize the training process and maintain the same loss scale as the conventional model, we normalize the quality scores by \textit{softmax}:
\begin{gather}
    \alpha_{j}^{i} = \frac{\exp (\hat{\alpha}_{j}^{i})}{\sum_{t=1}^{n} \exp (\hat{\alpha}_{t}^{i})}
    \label{eqa:8}
\end{gather}

\noindent \textbf{Sentence-Level Translation Quality (STQ)}: We take the average of the token hidden states as the sentence-level features, written as $\bm{h}_{x^{i}}$ and $\bm{h}_{\hat{y}^{i}}$. The sentence-level quality can be calculated as:
\begin{gather}
    u = \cos(\bm{h}_{x^{i}}, \bm{h}_{\hat{y}^{i}}) \\
    \hat{\beta}^{i} = u^{k}
    \label{eqa:9}
\end{gather}

% \begin{gather}
%     \hat{\beta}^{i} = \left( \frac{\bm{h}_{x^{i}} \cdot \bm{h}_{\hat{y}^{i}}}{||\bm{h}_{x^{i}}|| \cdot ||\bm{h}_{\hat{y}^{i}}||} \right)^{k}
%     \label{eqa:9}
% \end{gather}

Similarly, sentence-level quality scores are also normalized by \textit{softmax}:
\begin{gather}
    \beta^{i} = \frac{\exp (\hat{\beta}^{i})}{\sum_{t=1}^{M} \exp (\hat{\beta}^{t})}
    \label{eqa:10}
\end{gather}
where $M$ represents the batch size.

\subsubsection{JS-Divergence based Confidence Estimation (JS)}
An alternative of CP is \textit{Two-Pass} JS-divergence, which can reflect the difference between token distributions. It can be formulated as 
\begin{gather}
    JS(p||q) = \frac{1}{2}KL(p||r) + \frac{1}{2}KL(q||r)
\end{gather}
where $p$ and $q$ represent the distributions of tokens at each force-decoding step with different \textit{dropout}, and $r=(p+q)/2$.

\textbf{Token-Level JS Score} $\alpha_{j}^{i}$ is the JS score of $j$-th token in sentence $i$ during force-decoding. 

\textbf{Sentence-Level JS Score} $\beta^{i}$ is represented by the mean of token-level JS confidence in the $i$-th sentence.

Both of $\alpha_{j}^{i}$ and $\beta^{i}$ are multiplied by \textit{k} power and normalized by \textit{softmax}.

\subsubsection{Training Strategy}
Higher score indicates better quality. So the corresponding tokens or sentences should contribute more when computing loss, helping UNMT optimize in the reasonable direction. Therefore, we apply the quality scores to regulate the training loss. The loss of $i$-th sentence can be calculate as:
\begin{gather}
    \mathcal{L}_{i} = -\sum\nolimits_{j=1}^{n} \alpha_{j}^{i} \log P(x_{j}^{i}|\hat{y}^{i}, x_{<j}^{i}; \bm{\theta})
\end{gather}
And the total loss of mini-batch is:
\begin{gather}
    \mathcal{L} = \sum\nolimits_{i=1}^{M} \beta^{i} \mathcal{L}_{i}
\end{gather}

During the training, CP can be only employed in BT steps, while JS can be employed not only in BT steps but also AE steps because it actually measures the model confidence. In our experiments, JS and CP are respectively applied in AE steps and BT steps. Further analyses also compare the performance of different estimation methods.

\section{Datasets and Experiment Settings}
\subsection{Datasets}
\textbf{Pre-training:} For En-Fr, En-De, En-Ro, we download pre-trained language models from \texttt{XLM}\footnote{\url{https://github.com/facebookresearch/XLM}} and \texttt{MASS}\footnote{\url{https://github.com/microsoft/MASS}} toolkits.
For En-Zh, we train a standard XLM model from scratch. The monolingual data consists of WMT 2008-2019 News Crawl dataset (5M Chinese sentences in total and 5M English sentences uniformly selected for equality).

\noindent \textbf{UNMT:} For En-Fr, En-De, En-Ro, we respectively keep 2M (1M English, 1M the other language) sentences for training from WMT News Crawl. For En-Zh, we extract Chinese sentences from the first half of the 2M parallel sentences in LDC, and English sentences from the other half. WMT \textit{newstest 2013/2014}, \textit{newstest 2013/2016}, \textit{newsdev/newstest 2016} and NIST03/NIST06 as validation/test sets for En-Fr, En-De, En-Ro, and En-Zh, respectively.

%  For En-Fr, En-De, En-Ro, we respectively keep 2M (1M English, 1M the other language) sentences for training from WMT News Crawl. For En-Zh, we extract Chinese sentences from half of the 2M parallel sentences from LDC, and extract English sentences from the complementary half. WMT \textit{newstest} 2013/2014, \textit{newstest} 2013/2016, \textit{newsdev/newstest} 2016 and NIST03/NIST06 as validation/test sets for En-Fr, En-De, En-Ro, and En-Zh, respectively.

\subsection{Settings}
\textbf{CL Settings:} For difficulty computation, MUSE\footnote{\url{https://github.com/facebookresearch/MUSE}} is applied to map the monolingual word embeddings\footnote{we download fasttext embeddings pretrained on wiki, \url{https://fasttext.cc/}} into the common space. $c_{0}=0.01$ for En-De, En-Ro and En-Zh, $c_{0}=0.1$ for En-Fr. $T$ is approximately estimated by the step when UNMT baseline reaches 90\% BLEU \cite{papineni-etal-2002-bleu} on the valid set.

\noindent \textbf{UNMT Settings:} During training, mini-batches are limits to 2000 tokens and maximum sequence length is 100 tokens. Adam with $\beta_{1}=0.9, \beta_{2}=0.998, lr=0.0001$ is employed for optimization. When decoding, we use beam size as 4 and length penalty as 1.0 for each language pair. 4-gram BLEU score computed by \textit{multi-bleu.perl}\footnote{\url{https://github.com/moses-smt/mosesdecoder}} script is reported for comparison.

\section{Experimental Results}
\subsection{Translation Quality}
Table \ref{Tab.1} shows the UNMT results on different translation tasks. XLM and MASS are the baseline results \footnote{With the limitation of resources, the size of our training datasets is less than 2\% of the ones used in \cite{NEURIPS2019_c04c19c2}. Therefore, the baseline results are a bit lower.}. Our proposed method consistently outperforms the strong baselines, demonstrating the effectiveness of our method. Furthermore, removing either batch-level CL or sentence/word-level CL decreases the translation improvements on most language pairs, indicating the two parts are complementary. 
\begin{table*}[htpb]
\fontsize{10}{12}\selectfont
\centering
\begin{tabular}{lcccccccc}
\hline
\multirow{2}{*}{\textbf{Model}} & \multicolumn{2}{c}{\textbf{En-Fr}}       & \multicolumn{2}{c}{\textbf{En-De}}       & \multicolumn{2}{c}{\textbf{En-Ro}}       & \multicolumn{2}{c}{\textbf{En-Zh}}       \\
\cline{2-9} 
  & En→Fr   & Fr→En   & En→De    & De→En    & En→Ro     & Ro→En   & En→Zh   & Zh→En \\
\hline
XLM  & 35.89  & 33.58  & 26.21  & 32.51  & 33.48             & 30.97  & 12.97  & 26.42  \\
\ \ + Both Level & \textbf{36.31} & \textbf{33.97} & \textbf{27.22}      & \textbf{33.26} & \textbf{35.05} & \textbf{32.00}      & \textbf{13.70} & \textbf{28.18} \\
\cdashline{1-9}
 \ \ \ \ \ w/o \small{s/t level} & 35.70 & 33.77 & 26.27 & 32.69       & 34.04 & 31.78 & \textbf{13.70} & 27.33 \\
 \ \ \ \ \ w/o \small{batch level} & 35.91 & 33.90 & 27.01 & 33.21 & 34.72 & 31.58 & 13.30 & 27.04 \\
\hline \hline
MASS  & 34.97 & 32.98 & 26.93 & 32.20 & 34.32 & 31.58 & - & - \\
\ \ + Both Level & \textbf{35.36} & \textbf{33.40} & \textbf{27.53} & \textbf{32.62} & \textbf{34.86} & \textbf{32.27}      & - & - \\
\hline
\end{tabular}
\caption{\label{Tab.1}
BLEU scores of different UNMT methods for translations to and from English. Experiments on XLM are listed above the double lines and experiments on MASS are listed below it. "s/t" means "sentence/token".}
\end{table*}

Another interesting finding is that sentence/word level CL is more effective on similar languages, such as En-Fr, En-De, while single batch-level CL is suitable for the distant language pair like En-Zh. We assume that cross-lingual PLM on similar languages could provide hidden states with accurate semantic information, precisely estimating the quality of pseudo bi-text. In contrast, distant languages cannot fully take the advantage, while heuristic difficulty criteria help more. 

\subsection{Convergence Speed}
\begin{figure*}[htbp]
    \centering
  	\scalebox{0.60}{\includegraphics{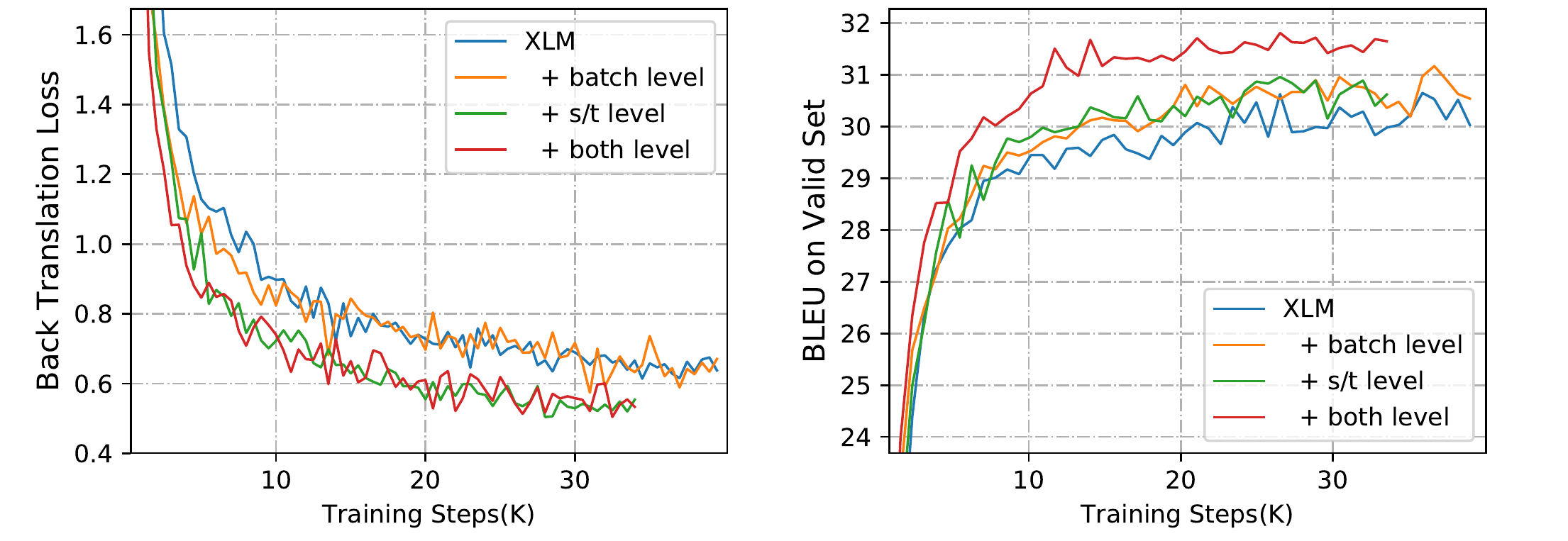}}
  	\caption{Average loss on the training process (left) and learning curves on valid data set (right) of WMT En-Ro . Our method achieves lower loss and higher BLEU score with faster convergence speed.}
  	\label{fig.3}
\end{figure*}
Most curriculum learning methods aim to accelerate convergence speed while improving performance. We visualize the average loss of training samples and the learning curve to compare the convergence speed on WMT Ro→En \textit{newstest2016} in Figure \ref{fig.3}. Both of the curves indicate that our method achieves convergence at a higher speed. 
\begin{table}
    \small
    \centering
    \begin{tabular}{ccc}
    \hline
    \multirow{2}{*}{\textbf{Language Direction}} & \multicolumn{2}{c}{\textbf{Our method}} \\
    \cline{2-3}
    & Step Acc.  & Time Acc. \\
    \hline
    En→De & 5.91x & 4.86x \\
    De→En & 2.46x & 1.95x \\
    En→Ro & 2.78x & 2.15x \\
    Ro→En & 3.08x & 2.41x \\
    \hline
    \end{tabular}
\caption{\label{Tab.2}
Acceleration on steps and time upon WMT En-De \textit{newstest2016} and WMT En-Ro \textit{newstest2016}. The acceleration is calculated by the ratio of the steps(time) when the baseline model reaches convergence to the steps(time) when our methods achieve equivalent translation quality.}
\end{table}

The left part of Figure \ref{fig.3} shows the loss curves. At the beginning of the training process, the average losses of different methods decrease with different speeds. However, the loss curves of the batch level CL and the baseline almost coincide at the end. When adding sentence/word level CL, the model achieves a lower loss than baseline, demonstrating the rationality of our weighted learning objective.

On the other hand, the learning curves, which are represented by the BLEU on valid set, clearly describe the efficiency of our method. As shown in the right part of Figure \ref{fig.3}, XLM baseline reaches convergence at step 31k, while our approach achieves the same performance at step 10k, indicating that our methods are 3.1 times faster.

The acceleration ratios for different languages are recorded in Table \ref{Tab.2}. Our methods significantly accelerate the training process. Considering the time exhausted in the computation of quality estimation, we also calculate the time acceleration. The records indicate that our methods can achieve equivalent performance with less training time. 

\section{Analysis}
\subsection{Correlation Between the Difficulty and Improvements}
\begin{figure}[htbp]
    \centering
  	\scalebox{0.40}{\includegraphics{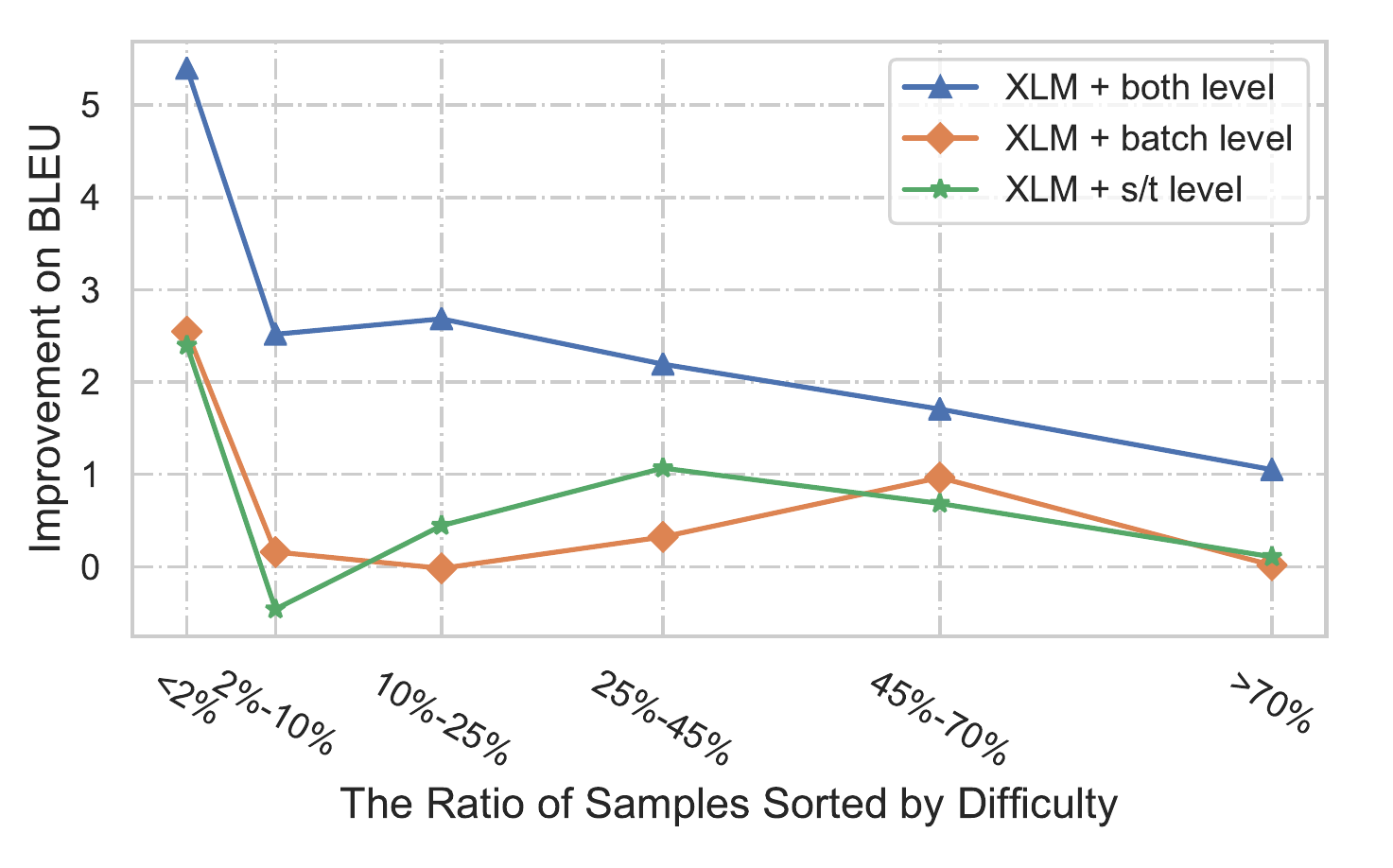}}
  	\caption{Improvements of BLEU at different difficulty intervals on WMT En→Ro \textit{newstest2016}.}
  	\label{fig.5}
\end{figure}

Even though our methods improve across all the language pairs, it remains a question which part of sentences contribute more to the performance. Figure \ref{fig.5} shows the BLEU improvements at different difficulty intervals on WMT En→Ro \textit{newstest2016}. The difficulty is represented by the definition described in section \ref{sec.3.1.1}. We find that our approach outperforms XLM baseline in different difficulty intervals. The easiest sentences (<2\%) have significant improvements, which owes to the emphasis on the easy samples during training. In contrast, hard sentences (>70\%) have limited performance gains. 

\begin{figure}[htbp]
    \centering
  	\scalebox{0.45}{\includegraphics{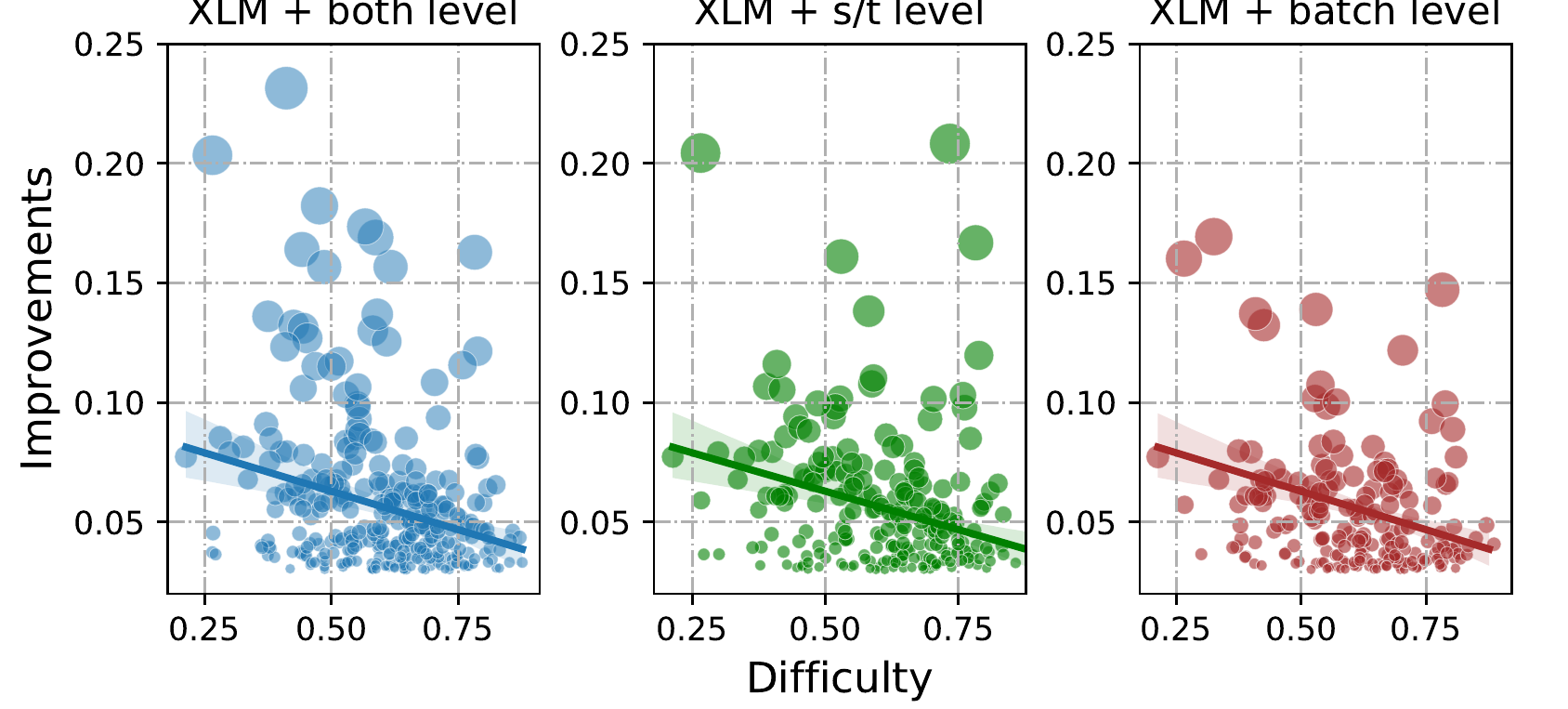}}
  	\caption{The relationship between difficulty distribution and improvements of BERTScore. The size of points indicate corresponding improvements.}
  	\label{fig.6}
\end{figure}

Figure \ref{fig.6} shows the relationship between the improvement of sentence-level BERTScore \cite{bert-score} and the difficulty distribution. The larger points are sparsely distributed on the left side, indicating that simple sentences achieve significant improvements. And the minor points are concentrated in the lower right corner, meaning that complex sentences yield slight performance improvement. 

This finding is different from related works on supervised NMT \cite{xu-etal-2020-dynamic,liu-etal-2020-norm}, which prove that curriculum learning is beneficial for complex samples. We suspect the reason lies in the particularity of our method, considering the quality of bi-texts instead of the pure difficulty. Therefore, we think our method helps the UNMT model mainly strengthen its essential translation ability.   

\subsection{Comparison of Difficulty Criteria}
To verify the effectiveness of our difficulty definition, we compare it with traditional difficulty criteria (such as length and rarity) on the single batch level CL. As shown in Table \ref{Tab.3}, the proposed definition achieves better performance than both length and rarity. By contrast, traditional artificial difficulties do not improve the UNMT translation quality and may even cause damage. 
\begin{table}
    \small
    \centering
    \begin{tabular}{lcccc}
    \hline
    \textbf{Method} & \textbf{De→En}  & $\Delta_{\text{BLEU}}$  & \textbf{Ro→En}  & $\Delta_{\text{BLEU}}$ \\
    \hline
    XLM    & 32.51  & -     & 30.97  & -  \\
    \cdashline{1-5}
    \ \ +$\text{BT}_{\text{Length}}$ & 32.21  & -0.30 & 31.04 & +0.07  \\
    \ \ +$\text{BT}_{\text{Rarity}}$ & 32.08  & -0.43 & 30.95 & -0.02 \\
    \ \ +$\text{BT}_{\text{Ours}}$     & \textbf{32.69}  & +\textbf{0.18} & \textbf{31.78} & +\textbf{0.81} \\
    \hline
    \end{tabular}
    \caption{\label{Tab.3}
    The comparison of different difficulty criteria on WMT De→En \textit{newstest2016} and WMT En→Ro \textit{newstest2016}. $\Delta_{\text{BLEU}}$ represents the performance increase or decrease compared with XLM baseline.}
\end{table}

We assume that traditional difficulty criteria are not appropriate for UNMT because many noises exist in pseudo bi-text during BT steps, which significantly changes the distribution of sentence-level difficulty. By comparison, our difficulty definition considers word-level translation difficulties. Intuitively, words with lower difficulties can be translated at higher quality, producing pseudo bi-text with fewer noises and easy to learn. Therefore, our difficulty definition implicitly describes the influence of noise, making the learning schedule more suitable for UNMT.

\subsection{Comparison of Different Estimation Methods}
We also compare the effect of our quality estimation approach with different confidence estimation methods. This part of the experiments is conducted without batch level CL for more evident results. Table \ref{Tab.4} shows that $\text{AE}_{\text{JS}}+\text{BT}_{\text{CP}}$ yields the best results among the methods, indicating the proposed estimation method is more engaging for UNMT. On the other hand, we find that single CP helps while single JS almost does not affect. Uncertainty-based model confidence $\text{AE}_{\text{VAR}}+\text{BT}_{\text{VAR}}$ (VAR is the abbreviation of variance)\footnote{Computing VAR for each token needs \textit{Q-Pass} forward computation with different dropout, \textit{Q} is set as 5 in the experiments.}, which is proven to be helpful in supervised NMT \cite{wang-etal-2019-improving-back,wan-etal-2020-self}, achieves only limited performance improvements in UNMT. Besides, the computation of VAR is time-consuming, slowing down the training efficiency by 54\% at each step.
\begin{table}
    \small
    \centering
    \begin{tabular}{lccc}
    \hline
    \textbf{Method} & \textbf{En→De} & \textbf{En→Ro} & \textbf{Speed} \\
    \hline
    \fontsize{8}{12.5} $\text{AE}_{\text{None}}+\text{BT}_{\text{None}}$    & 26.21 & 33.48 & 3183 (1.00x)  \\
    \cdashline{1-4}
    \fontsize{8}{12.5} $\text{AE}_{\text{JS}}+\text{BT}_{\text{JS}}$ & 26.28 & 33.72 & 2119 (0.67x) \\
    \fontsize{8}{12.5} $\text{AE}_{\text{VAR}}+\text{BT}_{\text{VAR}}$ & 26.51 & 33.38 & 1454 (0.46x) \\
    \fontsize{8}{12.5} $\text{AE}_{\text{None}}+\text{BT}_{\text{CP}}$ & 27.01 & 34.35 & 2961 (0.93x) \\
    \fontsize{8}{12.5} $\text{AE}_{\text{JS}}+\text{BT}_{\text{CP}}$ & \textbf{27.22} & \textbf{34.72} & 2475 (0.78x) \\
    \hline
    \end{tabular}
    \caption{\label{Tab.4} The comparison of different estimation methods on WMT En→De \textit{newstest2016} and WMT Ro→En \textit{newstest2016}. ST methods are listed below the dash line. Average speed (tokens/s) is measured on NVIDIA V100 and numbers in brackets is the fraction compared with XLM baseline.}
\end{table}

BT step is of vital importance for the translation ability of UNMT, which can be described as a rough imitation of NMT steps. Uncertainty-based confidence estimation is practical when bi-texts are pure. However, when the information provided by the particular bi-text is not equal, great deviation would be brought into the estimation of VAR or $\mathbb{E}$. By contrast, the quality of bi-text is much essential under this circumstance. We think that is the reason why CP yields higher translation performance.

\subsection{STQ Versus TTQ}
As we described in section \ref{sec.3.2.1}, fine-grained quality scores are estimated on the sentence-level (STQ) and the token-level (TTQ). We also compare their influence on translation performance. As shown in Table \ref{Tab.5}, both STQ and TTQ improve the translation quality on WMT \textit{newstest2016} En-De and WMT \textit{newstest2016} En-Ro. Interestingly, STQ outperforms TTQ. We suspect that cross-lingual PLM can estimate sentence-level quality more accurately than the token-level. Intuitively, the combination of STQ and TTQ achieves better results.
\begin{table}
\small
\centering
\begin{tabular}{lcccc}
\hline
\multirow{2}{*}{\textbf{Method}}    & \multicolumn{2}{c}{\textbf{En-De}}  & \multicolumn{2}{c}{\textbf{En-Ro}} \\
\cline{2-5}
& en→de  & de→en  & en→ro  & ro→en  \\
\hline
XLM  & 26.21  & 32.51  & 33.48  & 30.97 \\
$\text{AE}_{\text{JS}}+\text{BT}_{\text{CP}}$            & \textbf{27.22} & \textbf{33.26} & \textbf{34.72} & \textbf{31.58} \\
\cdashline{1-5}
\ \ \ \ \ \ w/o $\text{BT}_{\text{STQ}}$ & 26.69 & 32.54 &  34.42 & 31.34 \\
\ \ \ \ \ \ w/o $\text{BT}_{\text{TTQ}}$ & 26.92 & 32.88 & 34.59 & 31.43  \\
\hline
\end{tabular}
\caption{\label{Tab.5}
Comparison of sentence-level (STQ) and token-level (TTQ) quality estimation. STQ performs better than TTQ. }
\end{table}

\subsection{Effect of \textit{k}}
\begin{figure}[htpb]
    \centering
  	\scalebox{0.40}{\includegraphics{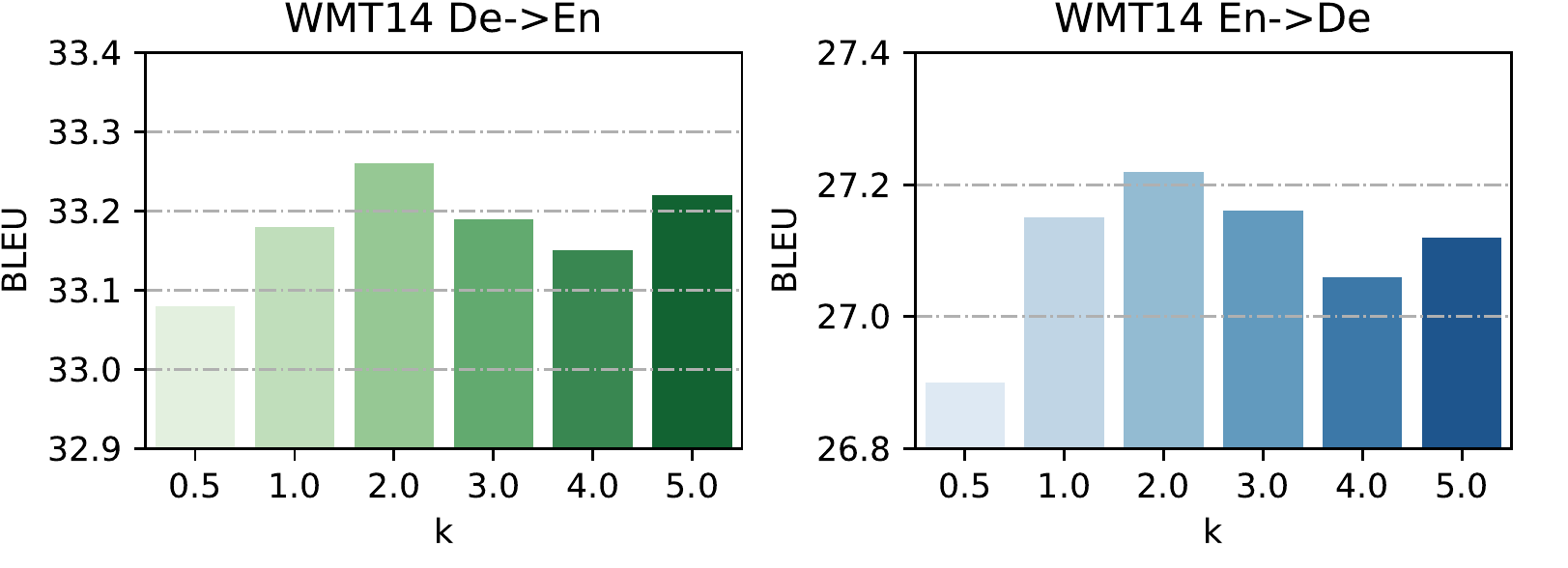}}
  	\caption{The effect of \textit{k} on the performance upon WMT En-De \textit{newstest2016}.}
  	\label{fig.4}
\end{figure}

Fine-grained quality scores in the proposed method are calculated by the \textit{k}-th power of cosine similarity. Therefore, we compare the translation performance with different \textit{k}. The results are shown in Figure \ref{fig.4}. Histogram illustrates that UNMT yields the best performance when $k=2$. However, when $k > 2$, the translation quality slightly decreases. We assume that appropriate $k$ can help our approach accurately reflect the translation quality, benefiting the UNMT performance.

% \subsection{Analysis on CT}
% \textbf{Stability on training size:} Curriculum learning is found to be sensitive to the hyper-parameters\cite{wan-etal-2020-self}. When the size of dataset changes, the hyper-parameters need to be modified. However, ST, which weights the loss rather than controlling the data loading process, is more stable. As shown in Figure \ref{fig.6}, XLM+ST gains almost equivalent improvements on different data sizes.
% \begin{figure}[htbp]
%     \centering
%   	\scalebox{0.40}{\includegraphics{figures/ST.pdf}}
%   	\caption{BLEU scores(\%) on WMT En-De \textit{newstest2014} with different training data sizes(using ST method only).}
%   	\label{fig.6}
% \end{figure}

\section{Related Work}
\subsection{UNMT}
\citet{lample2018unsupervised} and \citet{artetxe2018unsupervised} propose UNMT using monolingual corpus only, which established on the progress of cross-lingual word embedding projection \cite{artetxe-etal-2018-robust, lample2018word}. Recent years, UNMT rapidly develops with the help of pre-trained language models. \citet{NEURIPS2019_c04c19c2} releases the first cross-lingual PLM, named XLM, greatly improving the UNMT performance. \citet{song2019mass},\citet{liu-etal-2020-multilingual}, and \citet{NEURIPS2020_1763ea5a} designs different seq2seq pre-training strategy, achieving state-of-the-art UNMT performance. 

Even though various models are proposed, the key of UNMT is still the cross-lingual ability. \citet{sun-etal-2019-unsupervised} uses an agreement method to train UNMT with bilingual word embedding agreement. \citet{ren-etal-2019-explicit} ameliorates the cross-lingual ability of BERT \cite{devlin-etal-2019-bert} through predicting n-gram translation of masked tokens, benefiting the UNMT performance.  \citet{chronopoulou-etal-2020-reusing} modifies the predefined vocabulary of XLM for UNMT with limited monolingual corpus. 

However, most of previous work focuses the cross-lingual ability of word embedding or PLM but ignores the efficiency of the training process in UNMT.

\subsection{Curriculum Learning in NMT}
Curriculum learning \cite{10.1145/1553374.1553380} is motivated by the learning strategy of biological organisms which orders the training samples in an easy-to-hard manner \cite{elman1993learning}. 
It has recently shown its effectiveness on machine translation tasks by changing the order of training samples. \citet{kocmi-bojar-2017-curriculum} examine the effects of particular orderings of sentence pairs on the NMT training in one epoch. \citet{platanios-etal-2019-competence} propose competence-based curriculum learning framework, selecting samples at each step based on the difficulty and competence. \citet{liu-etal-2020-norm} use the norm of word embedding to modify the competence-based curriculum learning, improving the performance of supervised NMT. \citet{zhou-etal-2020-uncertainty} apply uncertainty into the difficulty and competence design. \citet{wan-etal-2020-self} adopt self-paced learning \cite{NIPS2010_e57c6b95} for NMT, replacing curriculum learning and yielding better performance. \citet{xu-etal-2020-dynamic} propose dynamic curriculum learning strategy for low-resource NMT. However, curriculum learning for UNMT is still unexploited and our work is the first attempt.

\section{Conclusion}
In this paper, we propose a multi-granularity CL method to improve UNMT. Specifically, a novel cross-lingual difficulty definition is first proposed to help UNMT learn from easy samples to the hard ones at batch level. Then, the qualities of pseudo bi-text at sentence/word-level are estimated by the model itself to regulate the loss function, automatically helping UNMT optimize in the appropriate direction. Empirical results show that our method outperforms the strong baselines on different language pairs with faster convergence speed. Further analyses confirm that our CL methods at different levels are helpful and complementary with each other, indicating the suitability for UNMT. In the future, we will explore its ability on multilingual machine translation and other cross-lingual generation tasks.

\section*{Acknowledgements}
The research work described in this paper has been supported by the Natural Science Foundation of China under Grant No. U1836221.

\bibliography{anthology,custom}
\bibliographystyle{acl_natbib}

\end{document}